%% file: IDE_Journal.tex
\newif\ifonecol

 \onecoltrue    

\ifonecol
\documentclass[twocolumn,technote]{IEEEtran}
\else
\documentclass{IEEEtran}
\fi
\usepackage{amsfonts}
\usepackage{amsmath}
\usepackage{amssymb}
\usepackage[final]{graphicx}
\usepackage{xspace}

\input{myheader}

\newcommand{\inst}[1]{\unskip${^{#1}}$}

\title{Fast Sparse Decomposition by Iterative Detection-Estimation}

\author{Arash Ali Amini,\inst{1}~\IEEEmembership{Student Member,}
        Massoud~Babaie-Zadeh,\inst{1}*~\IEEEmembership{Member,}
  \thanks{$^1$Electrical engineering department, Sharif university of
    technology, Tehran, Iran.}
  \thanks{$^2$Laboratoire des Images et des Signaux (LIS),
    Institut National Polytechnique de Grenoble (INPG), France.}
\thanks{This work has been partially funded by Sharif University of Technology, by French
Embassy in Tehran, and by Center for International Research and
Collaboration (ISMO).}
        and~Christian~Jutten \inst{1}~\IEEEmembership{Member} \\
  (Last update: November 2006)
  \thanks{Author's email addresses are: {\tt aaamini57@yahoo.com}, {\tt mbzadeh@yahoo.com} and {\tt Christian.Jutten@inpg.fr}}
  \ifonecol
  \thanks{Corresponding author: Massoud Babaie-Zadeh, email: {\tt mbzadeh@yahoo.com}, Tel: +98 21 66 16 59 25, Fax: +98 21 66 02 32 61.}
  \fi
  }

\begin{document}

\maketitle

\begin{abstract}
Finding sparse solutions of underdetermined systems of linear
equations is a fundamental problem in signal processing and
statistics which has become a subject of interest in recent
years. In general, these systems have infinitely many solutions.
However, it may be shown that sufficiently sparse solutions may be
identified uniquely. In other words, the corresponding linear
transformation will be invertible if we restrict its domain to
sufficiently sparse vectors. This property may be used, for
example, to solve the underdetermined Blind Source Separation
(BSS) problem, or to find sparse representation of a signal in an
`overcomplete' dictionary of 
primitive elements (i.e., the so-called atomic decomposition). The
main drawback of current methods of finding sparse solutions is
their computational complexity. In this paper, we will show that
by detecting `active' components of the (potential) solution, \ie
those components having a considerable value, a framework for fast
solution of the problem may be devised. The idea leads to a family
of algorithms, called `Iterative Detection-Estimation (IDE)',
which converge to the solution by successive detection and
estimation of
its active part. Comparing the performance of IDE(s) with one of
the most successful method to date, which is based on Linear
Programming (LP), an improvement in speed of about two to three
orders of magnitude is observed.

\end{abstract}

 \ifonecol

 \fi

\section{Introduction}\label{sec: intro}
Finding (sufficiently) sparse solutions of underdetermined
systems of linear equations has been studied extensively in recent
years~\cite{ChenDS99,CandRT06,DonoH01,DonoE03,EladB02,Fuch04,GribL06,GribN03,LiCA03,Trop04,ZibuP01}.
The problem has a growing range of applications in signal
processing. For example, it arises when dealing with
underdetermined sparse source
separation~\cite{GribL06,LiCA03,ZibuP01}. Another example is the
so-called `atomic decomposition' problem which aims at finding a
sparse representation for a signal in an overcomplete
dictionary~\cite{ChenDS99,CandRT06,Trop04}. Sparse representations
are more suited for content analysis, \ie extracting structure or
meaning from a signal. They may also be used to achieve
compression which in turn facilitates storage, processing and
communication of signals. Recently, interesting applications have
been reported in efficient (near-optimal) decoding of
`error-correcting codes'~\cite{RudeV05,CandTXX,CandT04}. Also,
some profound implications to the theory of sampling has been
found~\cite{DonoT06,Dono06}.

It is not surprising that this fundamental problem has such a wide
range of applications. The fact may simply be attributed to the
widespread use of linear systems and transforms (throughout
science and engineering). A linear transform (in a space with
finite dimension) may be represented by a system of linear
equations. Formerly, the underdetermined case, \ie the case of
`more unknowns than equations ', was considered degenerate and
undesirable due to non-uniqueness of the solution. In other words,
the corresponding linear transform is not invertible in this case
which greatly reduces its usefulness for modeling (real-world
problems). The general approach was (usually) to avoid the case by
reformulating the underlying (physical) problem (to obtain enough
equations in the unknowns). It is however possible to arrive at a
unique solution by imposing additional constraints. One such
constraint is (sufficient) sparsity of the solution, \ie to
require most components of the solution vector to be zero. More
specifically, it can be shown that for a (random) system with $n$
equations in $m (>n)$ unknowns, if there is a solution with less
than $n/2$ (out of $m$) nonzero components, then it is (almost
surely) the unique sparsest solution~\cite{Dono04}. In other
words, by limiting the domain of the underlying transform to
`sufficiently sparse' vectors, we can ensure its invertibility. We
may even take one step further and claim that sparsity is usually
more desirable than restrictive when it comes to signal processing
applications. For example, in the context of atomic decomposition,
a sparse solution leads to an efficient compact signal
representation.

On the other hand, recent theoretical results~\cite{Dono04}
provide a solid mathematical basis for some of the methods (and
optimality measures) experimentally found to produce sparse
solutions. But issues still remain, perhaps one of the most
important being the computational complexity of the available
methods~\cite{GribL06}. Our main objective in this paper is to
introduce a framework which may be used to achieve fast sparse
decomposition. But first, to get a better understanding of the
problem, we review two contexts in which the problem arises,
namely `Atomic Decomposition' and `Sparse Component Analysis
(SCA)'. Then we will review some of the available methods which
will be used as a basis for comparison when evaluating the
performance of our proposed method. We conclude the introduction
with a brief layout of the rest of the paper.

In the atomic decomposition viewpoint~\cite{MallZ93,ChenDS99}, we
have `one' signal whose samples are collected in the $n\times 1$
signal vector $\sbb$ and the objective is to express it as a
linear combination of a set of predetermined signals where their
samples are collected in vectors $\{\phib_i\}_{i=1}^{m}$. After
\cite{MallZ93}, the vectors $\phib_i$ are called \emph{atoms} and
the collection is called a \emph{dictionary}. In mathematical
language:
\begin{equation}\label{eq: atomic decomp expansion}
  \sbb = \sum_{i\,=1}^{m} \alpha_{i}\,\phib_{i} = \Phib \,\alphab
\end{equation}
where $\Phib$ is the $n\times m$ dictionary matrix (with columns
$\phib_i$) and $\alphab$ is the $m \times 1$ coefficient vector.
To represent any $n \times 1$ vector, a basis of $\Rn$ is
sufficient, \ie a collection of $n$ linearly independent vectors
(in $\Rn$). But if we take the number of atoms (much) more than it
is required ($m \gg n$), then the likelihood that a given signal
vector has a representation in terms of only a few (i.e. much less
than $n$) atoms is greatly increased. In that case, most of
coefficients in the expansion would be negligible, \ie the
coefficient vector would be sparse. In fact with proper selection
of dictionary, we may be able to find sparse representations for
most of the signals of a signal space of interest. As mentioned
before, such representations better reveal signal structure and
are highly desirable from a practical point of view. A dictionary
with $m > n$ atoms is called `overcomplete' and the corresponding
problem is usually referred to as `Atomic/Sparse
Decomposition'~\cite{ChenDS99}. It is clear that this problem is
essentially that of finding sparse solutions of an underdetermined
linear system.

In the SCA viewpoint~\cite{GribL06,LiCA03,BofiZ01,ZibuP01}, we use
the sparsity assumption to solve the so-called `underdetermined'
Blind Source Separation (BSS) problem~\cite{Card98,HeraJ86}. The
general BSS problem may be stated as: recovering $m$ unknown
source signals from $n$ known mixtures of them, when little
details are available about the sources and the mixing system. For
example, usually the only assumption (or information) about the
sources is their statistical independence. Similarly, regarding
the mixing system only general properties (such as
linearity/nonlinearity, being convolutive/instantaneous, ...) are
assumed. Here, we only consider the most common mixing model, i.e.
the (noiseless) linear instantaneous model:
\[
 \xb(t) = \Ab \,\sbb(t), \quad t = 1,\cdots, N
\]
where $\sbb(t)$ and $\xb(t)$ are the vectors containing sources
and mixtures and  $\Ab$ is the (unknown) $n \times m$ mixing
matrix. The only known quantity is $\xb(t)$. The objective is to
find the source vector and the mixing matrix only by observing
$\xb(t)$. For the case of `equal sources and mixtures' ($m=n$) and
with the assumption of an invertible mixing matrix $\Ab$,
estimation of $\Ab$ is sufficient to solve the problem. But in the
underdetermined case where the number of sources is more than
mixtures ($m > n$), even with the knowledge of $\Ab$, the system
is not invertible and we are unable to obtain the sources. As
mentioned before, this is where the added assumption of sparsity
is helpful. More specifically, \emph{if the original source vector
$\sbb$ is sufficiently sparse, then it is the unique sparsest
solution of $\xb = \Ab \sbb$}~\cite{Dono04}. Again, the problem
reduces to that of finding the sparse(st) solution of an
underdetermined system. It is also interesting to note that
`sparsity' may also be used to estimate $\Ab$, by applying
clustering techniques to the scatter plot of
$\xb(t)$~\cite{BabaJMC05,Hull99}. We, however, assume $\Ab$ to be
known (or estimated) a priori. Moreover, we assume that the
energy of the columns of $\Ab$ are normalized to 1, that is
$\|\ab_i\|_2=\ab_i^T \ab_i=1$ (this is always possible because as
it is seen in (\ref{eq: atomic decomp expansion}), each $\phib$
may be multiplied by a scalar and the corresponding coefficient
divided by that scalar. In BSS, this is usually called ``scale
indeterminacy''). It is also notable that sparsity is not much of
a restriction in practice: Many natural signals exhibit sparsity
either in the time-domain or in a transform-domain~\cite
{BofiZ01,GribL06,LiCA03}.

For future discussions, we will mainly adopt the terminology and
notation of SCA, although some references might be made to the
atomic decomposition terminology. This is partly because nearly
all the methods to be reviewed have been originally developed in
the context of atomic decomposition.

The methods used for sparse decomposition may be divided into two
categories: those selecting a solution of the underdetermined
system by \emph{minimizing a cost function} over the space of all
possible solutions, and those taking a more \emph{algorithmic
approach} without explicitly specifying a cost function. For the
methods of the first type, the cost function may be viewed as a
\emph{measure of sparsity}\footnote{To be more precise, the cost
function should be considered as a measure of deviation (or
departure) from sparsity, but for the sake of simplicity we will
neglect such formality.} of the solution vector. One such measure,
which is strongly supported by our intuition of sparsity (and may
even be considered its definition), is the so-called $l^0$~norm of
$\sbb$ denoted by $\norm{\sbb}{0}$ and defined as the number of
nonzero elements of $\sbb$. Unfortunately, minimizing the $l^0$
norm requires combinatorial search which quickly becomes
intractable as the dimension increases; It is also highly
sensitive to noise. It has been shown first
experimentally~\cite{ChenDS99} and then
theoretically~\cite{CandRT06,Dono04,DonoH01,DonoE03,EladB02,Fuch04,GribN03}
that the $l^0$~norm could be replaced by $l^1$~norm, \ie we seek a
solution minimizing $\norm{\sbb}{1} = \sum_{i=1}^{m} |s_{i}|$. The
$l^1$~norm is more robust to noise and more importantly, the
associated optimization problem is `convex' which can be solved
much more efficiently. The problem may also be stated as a Linear
Programming (LP) problem and then solved in polynomial time using
interior-point methods. Minimizing $l^1$ norm, which was initially
named Basis Pursuit (BP), may be considered the most successful
method to date. We will refer to this method as the `LP approach'
to emphasize that we will use linear programming techniques
(mostly interior-point solvers) to obtain its solution.

Besides LP, we also consider two other earlier approaches to
atomic decomposition. One of them, which we denote as the Method
of Frames (MOF) following~\cite{ChenDS99}, obtains a solution of
$\xb = \Ab \sbb$
having minimal $l^2$ norm, \ie $\norm{\sbb}{2}=(\sum_{i=1}^{m}
s_{i}^{2})^{1/2}$. The method has been originally developed
without any regard of sparsity~\cite{Daub88}, and it turns out
that its solution is usually not sparse. But merely as a method
of decomposition, it has some nice properties: the solution is
linear in $\xb$ and it may be obtained using the pseudo-inverse
of $\Ab$, \ie $\sbb_{\text{MOF}}=\Ab^{T}(\Ab\Ab^{T})^{-1}\xb$. It
may also be considered as the best linear inverse system in the
Least Squares (LS) sense (both statistically and
deterministically). We will mainly use MOF as a benchmark for the
speed of algorithms\footnote{Because of the existence of highly
efficient numerical algorithms for the computation of
pseudo-inverse (with computational cost close to solving a linear
system of comparable dimensions), MOF may be considered to
achieve fastest decomposition.}.

The other approach is Matching Pursuit (MP) developed by Mallat
and Zhang~\cite{MallZ93} (who also coined the name atomic
decomposition). It may be considered an algorithmic approach and
one of the first methods to target sparsity of the solution
(though implicitly). Recall that in the atomic decomposition we
seek a linear expansion of $\xb$ in terms of atoms $\phib_i$. MP
begins by finding the best single-atom approximation of $\xb$ in
the LS sense, \ie $\xb \approx \xbh_{1} = s_i \phib_{i}$ where
$s_i$ and $\phib_i$ are selected such that $\norm{\xb - s_i
\phib_{i}}{2}$ is minimized (over $1 \le i \le m$). This is
equivalent to finding the atom which best correlates with $\xb$,
\ie for which $|\xb^{T}\phib_{i}|$ is maximum. If the residue $\xb
- \xbh_1$ is small enough, the algorithm is terminated, otherwise
the same process is repeated for the residue. In other words, at
each step, MP finds the best single-atom approximation of the
residue. In this sense, it is a \emph{greedy algorithm}
(selecting the best choice given the current situation). We have
a good chance of obtaining a sparse representation if the
algorithm terminates early (\ie with a number of atoms much less
than $m$). However, as with any greedy algorithm, there are
situations in which an early mistake would lead to large
deviation from the optimal solution. We will discuss this issue
further in the experimental results section.

Among methods of decomposition available, the fast methods (e.g.
MP or MOF) usually don't produce accurate results, while LP which
is guaranteed to obtain the exact solution (asymptotically) will
become very computationally demanding at large dimensions. Our
proposed algorithm (or framework) is an attempt to keep accuracy
while approaching MP and MOF in speed. We begin with a general
introduction of the `Iterative Detection-Estimation (IDE)'
framework, followed by a detection-theoretic motivation for the
derivation of IDE algorithms. We then develop two versions of such
algorithms denoted as `IDE-s' and `IDE-x', followed by some
comments on the choice of parameters. We conclude with a
discussion of experimental results comparing the performance of
the proposed algorithms to existing methods.


\section{Iterative Detection-Estimation}
Perhaps one of the main obstacles to implementation of many
optimal methods of sparse decomposition is the inherent
`combinatorial search' required. The obstacle is overcome if we
could somehow detect which components of the (original) source
vector $\sbb$ are `active'. By active sources we mean those having
a \emph{considerable value}, as opposed to those being
\emph{nearly zero} and denoted as being `inactive'. The key idea
here is to detect (or determine) the `activity' status of each
source separately (\ie independently of all the other sources).
The total number of detections required would be $m$ which is
linear in the problem dimension.

The problem with this approach is that optimal detection of
`activity' of a source requires exact knowledge of the values of
other sources. Our solution is to use a suboptimal detector with
the exact values replaced by some previously known estimate (or an
initial guess). This rough detection may (surprisingly) be used to
obtain a better estimate of source vector which in turn may be
used to enhance the detection. By iteratively applying a
detection-step followed by an estimation-step\footnote{This step
may also be called approximation or projection step depending on
the approach we use to obtain the estimate.} we can hopefully get
progressively better estimates and get closer to the original
source vector, hence the name `Iterative Detection-Estimation
(IDE)'. This convergence will be justified by our experimental
results, although the theoretical convergence proof is a tricky
and open question.

\idescheme Fig.~\ref{fig:IDE_scheme} illustrates a schematic
diagram of the algorithm in its general form. In this figure, $k$
is the iteration index, $\sbb^{(k)}$ and $\sbb^{(k+1)}$ are
respectively current and next estimate of the source vector, and
$\mathcal{I}_{\alpha}$ is the set of indices of the sources
detected to be active\footnote{Subscript $\alpha$ is used to
designate quantities related to active sources. Similarly,
subscript $\iota$ is used for inactive sources.}.

We begin the discussion by giving a motivation for the detection
step based on a simple (statistical) model of sparsity. We then
give two versions of the estimation step leading to two versions
of IDE, namely IDE-s and IDE-x. Throughout the discussion, $\kact$
and $\kinact$ will be used to denote the number of sources
detected active and inactive, respectively. Also, throughout the
development of the algorithm, the term `active (inactive) sources'
usually means those sources `detected active (inactive)'. We
sometimes use it to refer to original `active (inactive) source'.
The distinction should be apparent from the context.
\section{Detection Step}
\subsection{motivation}
To provide motivation for the detection step, we first use a
Mixtures of Gaussians (MoG) to distinguish active/inactive states
of a sparse source. This provides us with a simple (intuitive)
model of sparsity. More specifically, let $\mypi{0}$ be the
probability of $s_{i}$ being \emph{inactive} ($\mypi{0}
\lessapprox\ 1$ to insure sparsity) . Then, the value of an
inactive source is modeled by $\gaussian{0}{\sigmat{0}}$, and an
active source by $\gaussian{0}{\sigmat{1}}$, where $\sigmat{0} \ll
\sigmat{1}$
 \footnote{A shorthand notation would be $s_i \sim
\mypi{0}\gaussian{0}{\sigmat{0}} +
(1-\mypi{0})\gaussian{0}{\sigmat{1}}$}. The probability $\mypi{0}$
will not be used in the development of the algorithm, but will be
useful as a measure of sparsity in the experimental results.

As stated previously, we detect the activity status of each source
separately. Assume that we want to determine the status of the
$i$-th source $s_i$. We observe $\xb = s_{i} \ab_{i} + \sum_{j\neq
i} s_j \ab_{j}$ and we wish to decide which of the following two
hypotheses has occurred :
\begin{align*}
H_{0}&: \quad s_{i} \sim \gaussian{0}{\sigmat{0}}, \\
H_{1}&: \quad s_{i} \sim \gaussian{0}{\sigmat{1}}.
\end{align*}
This is essentially a binary hypothesis testing
problem~\cite{Scha91}. It may be argued that $t_i = \ab_{i}^{T}
\xb$ contains all the information regarding the discrimination of
the two hypothesis, \ie it is a sufficient statistic for the
problem (given the value of all the other sources). Defining
$\mu_i \triangleq \sum_{j\neq i}^{m} s_{j} \, \ab_{i}^{\tp}
\ab_{j}$ and noting that $t_i = s_i + \mu_i$, we can reformulate
the problem in terms of the sufficient statistic as $H_{k}: t_i
\sim \gaussian{\mu}{\sigmat{k}}$ for $k=0,1$.

We approach the problem in the Neyman-Pearson framework,
considering $\{ s_{j} \}_{j\neq i}$ to be parameters (rather than
random variables). Also, we do not assign priors to the
hypotheses. The optimal test (in the NP sense) would then be a
likelihood ratio test, \ie one which compares the likelihood ratio
to a threshold. For the problem at hand the critical region of
this test may be written as
\[
\log \frac{\sigma_{0}}{\sigma_{1}} +
  \left( \frac{1}{2 \sigma^{2}_{0}} - \frac{1}{2 \sigma^{2}_{1}} \right)
  (t_i - \mu_i)^{2} > \tau
\]
or after absorbing known constants into the threshold as
\[
    |t_i-\mu_i| > \eps
\]
where $\epsilon$ is the new threshold.
Recalling the definition of $\mu_i$, it is observed that
implementing the optimal test for activity of $s_{i}$ requires
knowledge of all the other sources\footnote{ Note that because of
the dependence of the critical region on the value of $\mu_i$,
there is no Uniformly Most Powerful (UMP) test.}.
As mentioned before, our solution is to replace them with their
estimates (obtained from a previous iteration or from an initial
guess). The resulting sub-optimal test is then
\[
    g_{i}(\xb, \sbh) \triangleq \left| \ab_{i}^{\tp} \xb - \sum_{j \neq i}
\sh_{j} \, \ab_{i}^{\tp} \ab_{j} \right| > \eps
\]
for $1 \leq i \leq m$. We will call $g_{i}(\xb, \sbh)$ as defined
above the activity function associated with the $i$-th source.
Below, we have summarized the detection step where we have also
allowed the threshold to vary with iteration. It is found
experimentally that decreasing the threshold each iteration
produces better results.

\noindent \rule{\linewidth}{1pt} \\ \noindent \emph{Detection
Step} : Obtain active indices according to
\[
\mathcal{I}_{\alpha} = \{1 \leq i \leq m :  g_{i}(\xb,\sbh^{(k)})
> \epsilon^{(k+1)} \}
\]
\noindent \rule{\linewidth}{1pt} \\

\subsection{vector form}
It is possible to write the detection step in a simple form using
vector-matrix notations. Note that one may write the activity
function as
\begin{align*}
    g_{i}(\xb,\sbh) &= | \ab_{i}^{\tp} ( \xb \,
    - \, \Ab \sbh \, + \, \ab_{i} \sh_{i} ) | \\
                  &= | \ab_{i}^{\tp} ( \xb \,
     - \, \Ab \sbh) \, + \, \sh_{i} )|.
\end{align*}
If one collects the components $g_{i}(\xb,\sbh)$ in a `vector
activity function $\gb(\xb,\sbh)$', the detection step may simply
be stated as
\begin{align*}
    \gb(\xb,\sbh) = | \Ab^{\tp} ( \xb - \Ab \sbh) +  \sbh )| >
    \eps
\end{align*}
where $|\cdot|$ and $>$ operate component-wise when used on
vectors. Note that if the previous estimate $\sbh$ is itself a
solution of the system (i.e. $\xb = \Ab \sbh$), then the (vector)
activity function is simply $\gb(\xb,\sbh) = |\sbh|$ (this is the
case for IDE-s algorithm discussed below). But the previous
estimate does not need to satisfy the system, in which case the
term $\Ab^{\tp} ( \xb - \Ab \sbh)$ acts as a compensator (this is
the case for IDE-x). Also note that (as a special case) the
activity function evaluated at the true source vector is
$\gb(\xb,\sbb) = |\sbb|$. This is useful when selecting threshold
values.

\section{Estimation Step}
Knowing the sparsity pattern (i.e. active index set
$\mathcal{I}_{\alpha}$), the estimation of  sources would be
straightforward. Here, we introduce two simple approaches which
may be considered respectively as projections in the source space
(s-space) and the mixture space (x-space).
\subsection{s-space approach}
In this approach we obtain the source vector by solving the
following optimization problem:
\begin{equation}
    \sbh = \argmin_{\sbb} \sum_{i \in \, \mathcal{I}_\iota}
    s_{i}^{2} \;\; (\text{s.t.} \;\; \oursystem)
    \label{equ:optim1}
\end{equation}
where $\mathcal{I}_\iota = \mathcal{I}_\alpha^{c}$ is the inactive
index set. Let $\kact \triangleq |\mathcal{I}_\alpha|$ ($\kinact
\triangleq |\mathcal{I}_\iota|=m-\kact$) be the number of sources
detected active (inactive). The above operation may be thought of
as projection into the ($\kact$-dimensional) subspace determined
by the active indices. We denote the IDE algorithm using this
approach for source estimation as `IDE-s'.

Optimization problem~\eqref{equ:optim1} is a special case of
Quadratic Programming (QP) which has been extensively studied in
the literature~\cite{NocedalW99}. For simplicity, assume (for the
rest of this section) that `the first $\kinact$ sources' have been
detected inactive, i.e., $\mathcal{I}_\iota =
\{1,2,\cdots,\kinact\}$. Then, the cost function
in~\eqref{equ:optim1} may be stated as the quadratic form
$\sbb^{\tp} \Hb \sbb$
with $\Hb = \left( \begin{smallmatrix} \Ib_{\kinact} & \zerob \\
\zerob & \zerob \end{smallmatrix} \right)$ where $\Ib_{\kinact}$
is the $\kinact\times \kinact$ identity matrix.

Among the many numerically efficient approaches
available~\cite{NocedalW99,GouldHN01}, here we consider direct
solution of the so-called Karush-Kuhn-Tucker (KKT) system of
equations which serves as a necessary condition for
optimality~\cite{NocedalW99}, \ie the optimal solution should
satisfy
\begin{align*}
    \begin{pmatrix}
        \Hb & \Ab^{\tp} \\
        \Ab & \zerob
    \end{pmatrix}
    \begin{pmatrix}
        \sbb \\
        \lambdab
    \end{pmatrix} =
    \begin{pmatrix}
        \zerob \\
        \xb
    \end{pmatrix}
\end{align*}
where $\lambdab$ is the $n\times 1$ vector of Lagrange
multipliers. Under certain conditions, explicit formulas for the
solution of this system may be obtained. Partitioning vectors and
matrices into `inactive/active' parts, we have
\begin{align*}
    \begin{pmatrix}
        \Ib_{k} & \zerob &\Ainact^{\tp} \\
        \zerob & \zerob & \Aact^{\tp} \\
        \Ainact & \Aact  & \zerob
    \end{pmatrix}
    \begin{pmatrix}
        \sinact \\
        \sact \\
        \lambdab
    \end{pmatrix} =
    \begin{pmatrix}
        \zerob \\
        \zerob \\
        \xb
    \end{pmatrix}.
\end{align*}
Under the fairly general condition of $\kact \le n$, the `unique'
solution of the problem may be stated as
\begin{align}
\begin{cases}
\shinact = \Binact^{\tp} (\Binact \Binact^{\tp})^{-1} \Zb^{\tp}
\xb\\ \shact = (\Aact^{\tp} \Aact)^{-1} \Aact^{\tp}(\xb - \Ainact
\sinact)
\end{cases} \label{equ:optim1_sol1}
\end{align}
where $\Zb$ is a $n \times (n-\kact)$ matrix whose columns form a
basis for the null space of $\Aact^{\tp}$, and $\Binact \triangleq
\Zb^{\tp} \Ainact$. Another closed-form solution may be obtained
under a more restrictive condition, namely $\kact \le
\min\{n,m-n\}$. It can be shown~\cite{AminBJ06EUSIPCO} that in
this case, the `unique' solution of the problem is
\begin{equation}\label{equ:optim1_sol2}
    \begin{cases}
        \shact = (\Aact^{\tp} \Pb \Aact)^{-1} \Aact^{\tp}
                \Pb \xb \\
        \shinact = \Ainact^{\tp} \Pb \, (\xb -
            \Aact \sact)
    \end{cases}
\end{equation}
where $\Pb \triangleq (\Ainact \Ainact^{\tp})^{-1}$. Obtaining the
solution using these two explicit formulas is usually faster than
directly solving the $(m+n)\times(m+n)$ KKT system. In the
experiments of this paper, the second
closed-form~\eqref{equ:optim1_sol2} will be used.
\subsection{x-space approach}
In this approach, the source estimate is obtained as the solution
of the following optimization problem:
\begin{equation}
    \begin{cases}
        \sbh_{\alpha} = \argmin_{\sact} \norm{\xb-\Aact
        \sact}{2} \\
        \sbh_{\iota} = \zerob
    \end{cases}.
    \label{equ:optim2}
\end{equation}
In other words, we estimate the active part of the source vector
by projecting $\xb$ into the subspace spanned by the (allegedly)
active atoms, and simply set the inactive part to zero. Since this
expansion of $\xb$ in terms of the active atoms occurs in the
mixture space, we denote the associated IDE method as `IDE-x'.

Using pseudo-inverse of $\Aact$, the solution of
\eqref{equ:optim2} may simply be stated as (assuming $\kact \le
n$)
\begin{equation}
\begin{cases} \shact = (\Aact^{\tp} \Aact)^{-1} \Aact^{\tp}\xb \\
\shinact = \zerob
\end{cases}.
\label{equ:optim2_sol1}
\end{equation}
It is interesting to note that setting $\shinact$ to zero in
\eqref{equ:optim1_sol1} also leads to the same result. Since we
only care about true values of active sources and expect inactive
ones to be nearly zero, this is a reasonable simplification. In
this sense, IDE-x may be considered an approximation of IDE-s. It
is important to note that the IDE-x solution no longer satisfies
$\xb = \Ab \sbb$, and hence as later experiments show, this
slightly lowers the accuracy of IDE-x relative to IDE-s. The loss
is, however, negligible when the noise over the inactive part of
the solution is not high. This is the price we pay for \emph{the
tremendous gain in speed obtained due to the simplified structure
of the IDE-x estimate.}

\section{Initial Conditions}
To initiate iterations, we need an initial estimate. For all the
experiments in this paper, we will use the simplest initial
condition, i.e. $\sbh^{(0)} = \zerob$. Note that this is not a
solution of $\xb = \Ab \sbb$, but as was mentioned before, the
detection step does not require the (initial or middle) estimates
to satisfy the system. Also note that in the absence of prior
information, the `zero initial condition' is perhaps the most
reasonable one, because due to the sparse nature of the actual
solution, most sources would be zero anyway.

One may also use other `cheap' estimates initially. For example,
IDE may be used to improve upon the solution of the MOF method.

\section{Comments on the choice of thresholds}
In this section, we briefly discuss some issues regarding
threshold selection. First consider the ideal case where the
`actual' inactive sources are (exactly) zero. Now suppose that 1)
the detection step `at least' detects the actual active sources
correctly (there might also be some actually inactive ones,
incorrectly detected active). Then if 2) the solution of the
estimation step is unique, it will coincide with the actual sparse
solution since the latter achieves a cost function value of zero
(for both IDE-s and IDE-x). In other words, the estimation step
compensates for the mistakes made during detection and correctly
estimates `all' inactive sources to be zero. One way to guarantee
the uniqueness of the solution [for either of \eqref{equ:optim1}
or \eqref{equ:optim2}] is by keeping the number of sources
detected active below the number of mixtures (i.e. $\kact < n$).

The two conditions above suggest that there are \emph{implicit
bounds} on the value of threshold. It should be low enough to
guarantee that (nearly) all the actual active sources are detected
correctly. On the other hand, it should be high enough to keep the
number of those detected active below $n$. The above argument then
suggests that within those bounds a rough detection is sufficient
and will lead to the desired solution. In practice, for the
moderately difficult problem\footnote{A sparse decomposition
problem gets difficult when $n/m$  decreases or the actual
solution becomes less sparse.} those bounds provide enough gap for
us to easily select thresholds. As will be seen in the
experimental section, it may even be possible to obtain threshold
sequences which work well for `families' of problems. We will also
see that IDE is even robust to errors in detection of actual
active sources, in the sense that minor `missed detections' are
corrected through iteration.

There are also \emph{explicit bounds} on the threshold. Recall
that $g_i(\xb,\sbb) = |s_i|$. This suggests that any bound on the
the absolute value of the sources would translate (somewhat
directly) into a bound on the threshold. One might then restrict
the threshold to $0 < \eps < K\cdot\norm{\sbb}{\infty}$ where $K
\gtrapprox 1$ (values of $K$ greater than unity may be used to
account for estimation errors). For simplicity, in all the
experiments of this paper, we will assume that the original source
vector is normalized to unit $l^\infty$ norm (i.e.
$\norm{\sbb}{\infty} = 1$) and then select thresholds in the
interval $(0,1)$ (i.e. $K=1$). In real applications, one needs to
estimate $\norm{\sbb}{\infty}$. One simple approach is to take the
activity function at the first iteration as an estimate of source
absolute value. Thus if the `zero initial condition' is used one
gets the estimate
$\norm{\gb(\xb,\sbh^{(0)})}{\infty}=\norm{\Ab^{T}\xb}{\infty}$.

\section{Experimental Results}
In this section, we will examine the performance of the two
versions of IDE, \ie IDE-s and IDE-x, and compare them to some of
the available methods. This will be done by discussing the results
of five experiments detailing different aspects of IDE behavior. %

In all the experiments, the $\Ab$ matrix will be generated
randomly by drawing each of its $m$ columns from a uniform
distribution on the unit sphere in $\Rn$. We will use the Gaussian
mixture model discussed earlier to generate source vectors in the
first three experiments. A different source model will be used for
the last two experiments which will be explained later. In any
case, we always normalize the source vector so that
$\norm{\sbb}{\infty} = 1$. This limits the choice of thresholds to
the interval $(0,1)$.


We will use SNR as a measure of quality (or accuracy) of the
solution produced by an algorithm. To measure complexity, the
total CPU time required by the algorithm will be used (although
this is not an exact measure of complexity, it provides us a
rough estimation). Depending on the context, two different forms
of SNR will be considered. When dealing with a single realization
(or sample) of the system $\oursystem$, we usually use what may
be called `Spatial SNR (SSNR)', which is defined as
$\norm{\sbb}{2}^{2}/\norm{\sbb -\sbh}{2}^{2}$ where $\sbb$ and
$\sbh$ are respectively the original and the estimated source
vectors\footnote{Note that here we average over the source (or
spatial) index, on a single time sample.}. Since we are dealing
mostly with large systems (e.g. $m = 500, 1000$), this form of
averaging is justified. When working with many samples of the
system $\{\xb(t) = \Ab \sbb(t)\}_{t=1}^{N}$, we usually average
over time (index) obtaining `Temporal SNR' for each source, \ie
\[
\text{(Temporal) SNR}_{i} = \frac{\sum_{t\,=1}^{N} s_{i}^{2}(t)}
        {\sum_{t\,=1}^{N} [s_{i}(t)-\sh_{i}(t)]^{2} }, \quad 1\le
        i \le m.
\]
The context indicates which SNR definition is being used, and
hence, we often omit the `spatial' or `temporal' prefixes.

For the purpose of comparison, three of the available
decomposition methods, namely MOF, MP, and LP, will be considered.
The emphasis is on LP since this is the one guaranteed to obtain
the spars(est) solution. In all the experiments, unless explicitly
stated otherwise, the LP solution is obtained using MATLAB 7.0
implementation of an `interior-point' LP solver (called LIPSOL).
Also, all the CPU times are measured on a 2.4GHz P4 CPU under
MATLAB 7.0 environment.

\subsection{experiment 1 - evolution toward the solution}
\tabideprogress %
\subsubsection{a typical setting}
In this experiment, we will study the typical behavior of IDE by
considering a `single' realization of a system with dimensions
$m=1024$ and $n=\lfloor 0.4m \rfloor=409$. The source vector is
drawn from a Gaussian mixture with $\mypi{0}=0.9,\,
\sigma_{0}/\sigma_{1}=0.01$ and is normalized so that
$\norm{\sbb}{\infty} = 1$. In a single realization, the actual
number of active components in the source vector is more
important than the $\mypi{0}$ parameter (which somehow measures
sparsity `on the average'). In particular, for the (random) source
vector considered here, the number of sources with absolute values
over $0.01$ is obtained to be $105$. This is nearly equal to $n/4$
which signifies a relatively difficult problem (as will be
proposed by experiment 4).

Both versions of the IDE algorithm have been applied to the
problem. In either case, a total number of six iterations has been
used with threshold values $\eps =
0.3,\,0.2,\,0.1,\,0.05,\,0.02,\,0.01$. This sequence has been
found experimentally to produce results as accurate as those of
LP, for the problem family characterized by $(\mypi{0} = 0.9, n/m
= 0.4)$.

The results obtained at the end of each iteration are summarized
in Table~\ref{tab:ide_progress}. For each of the IDE-s and IDE-x,
the number of sources detected active ($\kact$), the elapsed CPU
time in seconds, and the (spatial) SNR, all obtained at the end of
each iteration have been recorded. Also,
Fig.~\ref{fig:idps_progress} provides a more visual account of
IDE-s progress toward the solution (the progress of IDE-x is
similar). Each plot in this figure shows the original and the
estimated source vectors after an iteration, respectively
designated by small black and large gray dots. The vectors are
plotted against the source index (\ie the plots are $s_{i}$ or
$\sh_{i}$ versus $i$). We have also identified sources detected to
be active after each iteration by drawing a small
square above them. %
\figidpsprogress %

Based on these results we can make the following observations: At
first, due to the low starting threshold value $(=0.3)$, the
number of sources detected active is more than necessary ($158 =
\kact > \text{actual \# act.} \approx 105$). The number, however,
satisfies the the uniqueness condition (of the estimation part)
$\kact < n$ which enables IDE(s) to start the iteration. Also
note from the figure that (for IDE-s) not all the actual active
sources are at first detected. The figure shows that \emph{the
initial guess for active sources is highly improved after the
second iteration} and this improvement continues (though more
gradually) until the algorithm converges to the original
solution. There are also (a few) sources correctly detected
active at first, wrongly discarded at a later iteration, but
eventually re-detected at final iterations. This shows the
self-correcting capability of IDE; A property that a greedy
algorithm such as MP does not possess.

Note that for IDE-s, the final number of sources detected active
is near the actual value. For IDE-x, final $\kact$ is higher, but
the final solution has the same quality ($\text{SNR} \approx
10^3\,\,\text{or}\,\,30\,\text{dB}$). This is in accordance with
our previous statement that false alarm in detection of active
sources does not affect the performance as long as it remains
within the limits of the uniqueness condition.

Another notable observation is that for both versions, SNR
increases by nearly an order of magnitude every two iterations
until it reaches the final value of $\approx 10^3$ which as we
will see is comparable to the quality obtainable by LP. Also note
that \emph{each iteration of IDE-x is nearly an order of magnitude
faster than that of the IDE-s}; A property that holds in general
as will be confirmed in a later experiment.

\subsubsection{comparison of algorithms}
\tabalgorithmtimes %
In Table~\ref{tab:algorithm_times}, we have summarized the results
obtained by some of the available methods  when applied to the
same realization of the problem (along with those of IDE's). For
LP, both the interior-point and Simplex implementations are
considered. For MP, the results after 10, 100, and 1000 iterations
are recorded separately.

It is observed that both versions of IDE achieve a final SNR of
nearly $30$ dB (after six iterations) which is slightly better
than the $26$ dB obtained by LP. The major difference is in the
time required by each algorithm. In fact, with nearly the same
final SNR, the time comparison would be more meaningful.

\emph{We observe that IDE-x is ten times faster than IDE-s which
itself is a hundred times faster than LP-interior which in turn is
ten times faster than LP-Simplex}.
Thus, IDE-x, for example, achieves nearly four orders of magnitude
improvement in speed over LP-Simplex, which is a truly remarkable
achievement. The average results are more or less the same, as
will be discussed in the third experiment.

A comparison with the results obtained by MOF shows that it has
nearly the same speed as IDE-x. The final quality achieved
($\approx 2$ dB) is however far from acceptable. This is not
surprising since MOF was not meant originally to select the
spars(est) solution.

The quality and time obtained by MP after 10 iterations is very
close to those of MOF. The best performance is achieved around 100
iterations with a final SNR value of nearly $11$ dB and a time
comparable to that of IDE-s. This is the maximum quality
attainable by MP. It may partly be explained by recalling that in
the present problem, the number of (actual) active sources is
nearly 105 and that for MP, the number of (active) atoms present
in the expansion (of $\xb$) is the same as the number of
iterations. The claim is further confirmed by noting that after
1000 iterations the quality actually degrades to $\approx 10$ dB.
The observation reveals the fundamental problem of `greedy
algorithms' of which MP is one. We will discuss the problem
shortly and show how IDE-x effectively evades it.

\subsubsection{IDE-x versus MP}
Before concluding this experiment, we want to briefly comment on
how IDE-x may be used to improve upon MP. There is a resemblance
between the two algorithms. Recall that, at each step,  MP finds
the atom that best correlates with the residue (up to that point).
In this sense, MP finds successive `single-atom approximations' to
$\xb$ which at the end add up to be build the final estimate. In
contrast, at each iteration, IDE-x expands $\xb$ over all the
atoms detected to be active, and hence, it is more likely to
obtain the optimal (sparse) expansion.

Fig~\ref{fig:mp_vs_idpx} shows that this is indeed the case. In
this figure, the relative approximation error in the expansion of
$\xb$ is plotted versus iteration (or step) for both IDE-x and MP.
Note that MP requires nearly 1000 steps to achieve the same error
that IDE-x has achieved in 6 iterations. Moreover, in doing so, MP
incorporates into the expansion nearly all the 1024 atoms
available (recall that for MP each step adds one atom).
Consequently, the resulting $\sbb$ vector is far from sparse. This
reflects the main problem of greedy algorithms: making an early
mistake usually takes many steps to correct, during which the
algorithm deviates considerably form the optimal solution. IDE-x
(and in general IDE's) avoid this by expanding over all possible
candidates at each iteration.
\figmpvsidpx



\subsection{experiment 2 - average quality}
In this experiment, we compare average behavior of IDE's with that
of LP. The three algorithms are applied to $N = 1000$ time samples
$\{\xb(t)\}_{i=1}^{N}=\{\Ab\sbb(t)\}_{i=1}^{N}$. The `temporal
SNR' is then obtained for each algorithm and plotted against the
source index (i.e. (Temporal) SNR$_i$ versus $i$).
Fig.~\ref{fig:tempo_avg} shows the results for three illustrative
cases.%
\tempoavg%
For all the cases a Gaussian mixture model with
$\mypi{0} = 0.9,\, \sigma_{0}/\sigma_{1} = 0.01$ is used to
generate the $N$ time samples. The three plots correspond to
different choices of $(m,n/m)$ pairs, i.e. $(100,0.6)$,
$(500,0.6)$ and $(500.0.4)$ respectively.

A fixed threshold sequence, namely $\eps = 0.7,\,0.6,\,0.5,\,$
$0.4,\,0.3,\,0.2,\,0.1,\,0.07,\,0.05,\,0.02$, is used in all the
three cases and over all the $N$ samples. This sequence is found
(experimentally) to produce slightly better results than LP in all
cases of interest. \emph{Note that although we have set the
thresholds manually, they are only set once at the beginning and
there is no need to change them on a per-sample basis}. Also more
experiments with other combinations of the problem parameters
(i.e. $(m,n/m,\mypi{0})$) showed that this is indeed a `good'
choice for nearly all problems for which LP is `good', especially
at higher dimensions (i.e. for large $m$).

The three cases in Fig.~\ref{fig:tempo_avg} were chosen to
illustrate some general trends. Note that IDE's outperform LP as
shown by the gap between their average (temporal) SNRs, but the
gap reduces as the dimension is increased (i.e. increasing $m$
while $n/m$ is fixed). \emph{In other words, the performance of
the algorithms converges to one another as we increase $m$}. This
is confirmed by more experiments. Another trend is that the gap is
usually reduced \emph{as the problem gets harder} (i.e. decreasing
$n/m$ while $m$ is fixed). The third plot also shows that
surprisingly sometimes IDE-x (slightly) outperforms \mbox{IDE-s}.

\subsection{experiment 3 - average complexity}
In this experiment, we will examine the relative complexity (or
speed) of the algorithms more closely. The measure to be used is
the `average CPU time' required by each algorithm. More
specifically, we are interested in `average time' versus `problem
dimension' plots where the dimension is $m$, the number of
sources.
We select seven points in the interval%
\footnote{The points are selected to be equidistant in the
logarithmic scale, \ie $m=10,\,20,\,50,\cdots$} %
$10 \le m \le 10^{3}$, and for each $m$, we generate $N=10$
instances of the problem, keeping $n/m$ fixed at nearly $0.6$ (or
more exactly $n = \lfloor 0.6m\rfloor$). Each of the algorithms
under study is then applied to the $N$ samples and the average
time (obtained over the $N$ samples) is used as an index of
complexity at the specified dimension. Fig.~\ref{fig:speed_test}
summarizes the results.

\speedtest %

To generate the figure, all the iterative algorithms (\ie IDE-s,
IDE-x and MP) have been applied only for 10 iterations. Moreover,
we have only considered the interior-point implementation of LP.

Examining the figure, similar patterns as those encountered
earlier may be identified. Again, the slowest algorithm is LP
followed by IDP-s which is more than one order of magnitude
faster; The difference being nearly constant across dimension. It
is interesting to note that IDE-x may be grouped along with MP and
MOF as the fastest algorithms. The three algorithms have nearly
the same complexity at higher dimensions (\eg at $m=1000$).
\emph{We may then use IDE-x to achieve qualities near that of LP,
while keeping the complexity as low as those of MOF and MP.} Even
with IDE-s the speed improvement is considerable.

\subsection{experiment 4 - practical thresholds on sparsity}
As stated in Section~\ref{sec: intro}, to ensure uniqueness of
the sparsest solution, the number of active sources should be
limited to $n/2$. But in practice, most methods breakdown before
reaching this theoretical bound. In this experiment, we study
practical limits (on the number of active sources) for IDE-s,
IDE-x and LP.

In order to have more control over the sparsity,
we generate source vectors according to a different model other
than the Gaussian mixture. More specifically, given the number of
active sources, \nact, a source vector is generated with exactly
\nact\ of its components randomly selected to be unity. The rest
of the components, which represent inactive sources, are drawn
from a zero-mean Gaussian with variance 0.01. This allows for a
more accurate control of the sparsity. In fact, for this type of
source, the quantity $\#act/(n/2)$ acts as a (normalized) measure
of sparsity\footnote{Again to be accurate, the quantity should be
considred a measure of non-sparsity. To simplify discussion,
however, we neglect these technicalities.} very useful to our
discussion. Note that to ensure the `uniqueness of the sparsest
solution' property, $\#act/(n/2)$ should be kept below unity.

We will take $m=1000,\,n=400$ and select 25 values of
$\#act/(n/2)$ in the range $[0.1,1]$. For each $\#act$, both IDE's
and LP are applied to $N=10$ realizations of the problem and the
average SNR (over the $N$ samples) obtained by each method is
determined. Figure~\ref{fig:sparsity_thresh}(a) illustrates the
results when the general threshold sequence of experiment~2 has
been used for both
IDE's. %
\sparsitythresh%

Examining the figure, it is observed that the output SNR of both
IDE-s and IDE-x is increased monotonically up to $\#act/(n/2) =
1/2$, after which it descends steeply\footnote{Some of the
steepness is due to how the IDE's have been implemented...}
reaching nearly $0$ dB around $\#act/(n/2) = 3/5$. The behavior of
LP is somewhat similar except that the SNR begins to fall earlier
and the degradation is more gradual. In particular, LP's
performance is still acceptable around $\#act/(n/2) = 3/5$. A
general point to be made is that for all the three algorithms,
\emph{there seems to be thresholds on sparsity up to which they
perform well and after which they degrade quickly in quality.}

It is possible to enhance the performance of IDE near the sparsity
threshold by applying more iterations. To show this, we will
examine the behavior using a longer threshold sequence with values
spread wider across the $(0,1)$ interval. The specific values
are: $\eps = 0.9, 0.8, 0.7, 0.6, 0.5, 0.4, 0.3, 0.2, 0.1, 0.07,
0.05, 0.02, 0.01$. Figure~\ref{fig:sparsity_thresh}(b) illustrates
the results using this new threshold sequence. Note how IDE
performance now degrades more gradually after $\#act/(n/2) =
1/2$, keeping the SNR at an acceptable level around $\#act/(n/2)
= 3/5$; A behavior bearing more resemblance to LP.

Another interesting observation may be made by comparing the
high-sparsity (\ie low $\#act$) parts of the plots in
Fig.~\ref{fig:sparsity_thresh}(a) and (b): These parts are
essentially unaffected by changing the threshold sequence. This
result is in accordance with our previous intuitions. To sum up,
\emph{for relatively easy (\ie highly sparse) problems, IDE is not
sensitive to the choice of thresholds; Roughly general threshold
sequences may be used without sacrificing performance; It is for
difficult problems near the sparsity edge that the choice of
threshold sequence really matters. In fact, the sparsity (edge)
above which the method works is set by the chosen sequence.}

The observation we made that there is a threshold on $\#act$
(below the one suggested by theory) which limits the performance
in practice has been pointed out by various authors. In fact, the
figures we encountered for $\#act$ has also been obtained for the
LP approach before. For example, \cite{Dono04} reported the
experimental bound of $3n/10$ on $\#act$ for the minimum $l^{1}$
norm solution to coincide with the sparsest solution. The bound
$n/4$ has been obtained for the incomplete Fourier dictionary
in~\cite{CandRT06}. It appears that developing methods to fill the
gap and work right up to the $n/2$ limit would be one of the
challenges to be faced in the future.

\subsection{experiment 5 - sensitivity to noise in the matrix}
In SCA applications, where the $\Ab$ matrix is estimated from
mixture data, the robustness of the source-determination
algorithms to `estimation noise in $\Ab$' is important. This is
not the case for applications like atomic decomposition where the
dictionary $\Ab$ is pre-determined. Even in these cases some noise
may be induced on $\Ab$, for example, as a result of quantization.
In this experiment, we will examine the effect of these
perturbations on the performance of IDE's and LP.

To model the perturbations, we will add to every component of the
original matrix $\Ab$, a Gaussian noise of variance $\sigma_{A}
\times \max |a_{ij}|$. The columns of $\Ab$ are then re-normalized
to unit $l^{2}$ norm\footnote{The results were observed to be
nearly the same without normalization.}. To conduct the
experiment, we take a random source vector $\sbb$ with $n/8$ of
its components active (generated according to experiment 5 model),
a random $500 \times 200$ matrix $\Ab$, and 10 values for
$\sigma_{A}$ in the interval $[0.001,0.1]$. For each $\sigma_{A}$,
we generate $N=10$ noisy realizations
$\{\Abh_{k}(\sigma_{A})\}_{k=1}^{N}$ according to the procedure
mentioned above. An algorithm is then applied to the $N$ noisy
problems, designated with $\{(\sbb,\Abh_{k})\}$, resulting in the
estimated source vectors $\{\sbh_{k}(\sigma_{A})\}_{k=1}^{N}$.
Finally, the average (spatial) SNR in $\sbb$, \ie ${\small (1/N)
\sum_{k=1}^{N} \norm{\sbb}{2}^{2}/\norm{\sbb -
\sbh_{k}(\sigma_{\!A})}{2}^{2}}$, is plotted against the average
SNR in $\Ab$, defined as ${\small (1/N) \sum_{k=1}^{N}
\norm{\Ab}{F}^{2}/\norm{\Ab - \Abh_{k}(\sigma_{\!A})}{F}^{2}}$
where $\norm{\cdot}{F}$ denotes the Frobenius matrix norm.

\noiseinA%
The results are illustrated in Fig.~\ref{fig:noise_in_A} for the
algorithms IDE-s, IDE-x and LP. For IDE's, the general sequence of
experiment~2 has been used. A typical behavior is observed for the
three algorithms: They resist small amounts of noise in $\Ab$ (up
to SNRs of nearly 30 dB), but they degrade quickly in quality as
the noise is increased beyond some limit. Also note that \emph{the
quality gain of IDE's over LP is only obtained for very low-noise
$\Ab$ matrices}. The SNR curves for the three algorithms converge
as a result of an increase in $\Ab$-noise, indicating the loss of
performance gain. Another notable observation is that, at high
noise levels, IDE-x performs slightly better than both LP and
IDE-s which is somehow suggestive of a `de-noising' property. It
may be attributed to the fact that IDE-x seeks to minimize the
distance $\norm{\xb - \Abh \sbh}{2}$ unlike IDE-s and LP which
enforce $\xb = \Abh \sbh$ on the solution; An equation that need
not hold in the noisy cases.

\section{Conclusion}
We have shown that by (rough) detection of active sources, one can
eliminate the need for a combinatorial search, effectively
replacing it with one `comparison of an activity function against
a threshold' for each source. A possible choice for the activity
function $g_i(\xb,\sbh^{(k)})$ was proposed based on ideas from
binary hypothesis testing under Gaussian mixture prior for
sources. The detection step required an estimate of the source
vector, and together with an estimation step, it was used in an
iterative setting to obtain the `Iterative Detection-Estimation'
family of algorithms. We proposed two approaches for source
estimation (given that the sparsity pattern is roughly known): one
was based on projection of the solution set of $\xb = \Ab \sbb$
into the activity subspace in the `source space' leading to the
IDE-s algorithm. The other one was based on projection of $\xb$ on
the  subspace spanned by active atoms in the `mixture space' which
lead to  the IDE-x algorithm.

We showed experimentally that with proper threshold selection,
both versions of IDE can achieve accuracies comparable to LP (or
even slightly better) after few iterations. The interesting point
was that IDE's achieve this much faster, with IDE-s (IDE-x) being
nearly two (three) orders of magnitude faster than LP.

It was also observed that the algorithm is usually not `too
sensitive' to threshold values. In particular, a fixed threshold
sequence may be used for every instance of a fixed problem family
(determined by a fixed sparsity level and fixed $n/m$ value),
i.e., there is no need to modify the thresholds on a per-sample
basis. Also, a threshold sequence was found experimentally that
could be used over a wide range of problem families to produce
`good' results.

In general, these results suggest that IDE's might be used as fast
alternatives to LP when dealing with high-dimensional sparse
decomposition problems. One might also think of IDE as a general
framework of which the proposed algorithms are just two examples:
There might be better ways of detecting (single) source activity,
e.g.~using better activity functions, thresholdless decisions
(see below), etc. Similarly, there might be better implementations
of the estimation step, e.g. using different cost functions.

For example, one may think about a thresholdless variant of IDE:
we know from the uniqueness condition (Section~\ref{sec: intro})
that at most $n/2$ of sources may be active. Then, instead of
using thresholds on the values of the activity function for
detecting active sources, {\em all $n/2$ sources for which the
values of the activity function are the highest are detected to be
active}. Although using this approach no threshold is required,
it makes the algorithm somehow `greedy' (but of course not as
greedy as MP). Consequently, the algorithm may get trapped in
`local minima', specially where the degree of sparsity decreases
(this is verified by our first simulations). However, having no
thresholds is advantageous enough to use such a version in some
practical applications.


\bibliographystyle{IEEEtran}
\bibliography{SepSrc2}

\end{document}

%% file: myheader.tex
\newcommand{\sigmat}[1]{{\sigma}_{#1}^{2}}  
\newcommand{\xb}{{\bf x}}
\newcommand{\Ab}{{\bf A}}
\newcommand{\Bb}{{\bf B}}
\newcommand{\Zb}{{\bf Z}}
\newcommand{\Pb}{{\bf P}}
\newcommand{\Hb}{{\bf H}}
\newcommand{\Ib}{{\bf I}}
\newcommand{\zerob}{{\bf 0}}
\newcommand{\Phib}{{\bf \Phi}}

\newcommand{\sbb}{{\bf s}}

\newcommand{\ab}{{\bf a}}
\newcommand{\gb}{{\bf g}}
\newcommand{\alphab}{{\mbox{\boldmath $\alpha$}}}

\newcommand{\phib}{{\mbox{\boldmath $\phi$}}}

\newcommand{\mypi}[1]{{\pi}_{#1}}
\newcommand{\gaussian}[2]{\mathcal{N}(#1,#2)}

\newcommand\eg{e.g.,\xspace}
\newcommand\ie{i.e.,\xspace}
\newcommand{\oursystem}{\xb = \Ab \sbb}
\newcommand{\Rn}{\mathbb{R}^{n}}

\newcommand{\norm}[2]{\|#1\|_{#2}}

\newcommand{\lambdab}{{\mbox{\boldmath $\lambda$}}}
\newcommand{\sbh}{\hat{\sbb}}
\newcommand{\sh}{\hat{s}}
\newcommand{\Abh}{\hat{\Ab}}

\newcommand{\xbh}{\hat{\xb}}

\newcommand{\Aact}{\Ab_{\alpha}}

\newcommand{\Ainact}{\Ab_{\iota}}
\newcommand{\Binact}{\Bb_{\iota}}
\newcommand{\sact}{\sbb_{\alpha}}
\newcommand{\shact}{\sbh_{\alpha}}
\newcommand{\sinact}{\sbb_{\iota}}
\newcommand{\shinact}{\sbh_{\iota}}

\newcommand{\kact}{k_{\alpha}}
\newcommand{\kinact}{k_{\iota}}

\DeclareMathOperator*{\argmin}{arg\,min}
\newcommand{\tp}{T}
\newcommand{\eps}{\epsilon}
\newcommand{\nact}{$\#act$}

\newcommand{\tabideprogress}{
\begin{table}[tb]
\caption{IDE progress toward final solution} \centering
\begin{tabular}{|c | c || c | c | c || c | c | c|}
\cline{3-5}\cline{6-8}
 \multicolumn{2}{c||}{} & \multicolumn{3}{c||}{IDE-s} & \multicolumn{3}{c|}{IDE-x} \\
 \cline{1-2}\cline{3-5}\cline{6-8}
  $k$ &  $\eps^{(k)}$ & $\kact$ & $\Delta T$ & SNR
                      & $\kact$ & $\Delta T$ & SNR \\
\hline
        1  &  0.3   &  158    &  0.377  & 6.44
                    &  158    &  0.025  & 5.50 \\
        2  &  0.2   &  47     &  0.297  & 8.24
                    &  49     &  0.008  & 8.24 \\
        3  &  0.1   &  58     &  0.292  & 11.85
                    &  149    &  0.019  & 14.51 \\
        4  &  0.05  &  73     &  0.293  & 18.26
                    &  96     &  0.013  & 21.06 \\
        5  &  0.02  &  105    &  0.310  & 25.36
                    &  176    &  0.026  & 27.88 \\
        6  &  0.01  &  107    &  0.315  & 30.27
                    &  126    &  0.021  & 28.80 \\
\hline
\end{tabular}
  \label{tab:ide_progress}
\end{table}
}

\newcommand{\tabalgorithmtimes}{
\begin{table}[ptb]
\caption{Comparison of different algorithms} \centering
\begin{tabular}{|c || c | c|}
\hline
 algorithm         & total CPU time   &   SNR (dB) \\
\hline\hline
 IDE-s (6 itrs.)   & $1.88\,e\,00$    &  30.27\\
 IDE-x (6 itrs.)   & $1.12\,e\,{-1}$  &  28.80\\\hline
 LP (interior-pt)  & $1.23\,e\,{+2}$  &  26.25\\
 LP (simplex)      & $5.45\,e\,{+3}$  &  26.25\\\hline
 MP (10 itrs.)     & $1.54\,e\,{-1}$  &  1.80\\
 MP (100 itrs.)    & $1.58\,e\,{00}$  &  10.70\\
 MP (1000 itrs.)   & $8.71\,e\,{00}$  &  9.82\\\hline
 MOF               & $1.38\,e\,{-1}$  &  2.36\\
\hline
\end{tabular}
  \label{tab:algorithm_times}
\end{table}
}

\newcommand{\figidpsprogress}{
\begin{figure*}[ptb]
\centering
\includegraphics[width=\textwidth]{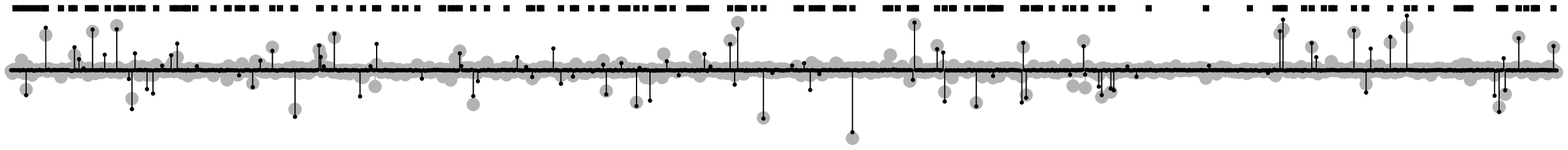}
\includegraphics[width=\textwidth]{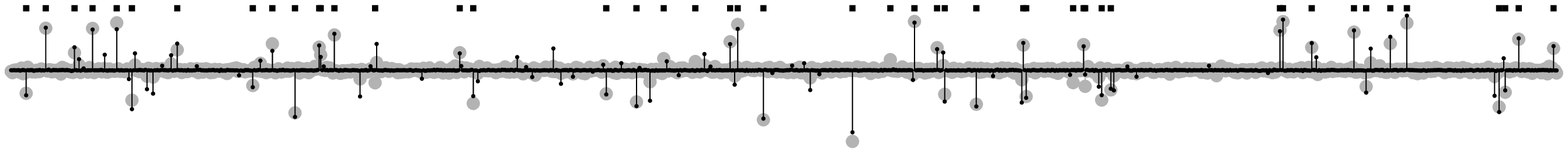}
\includegraphics[width=\textwidth]{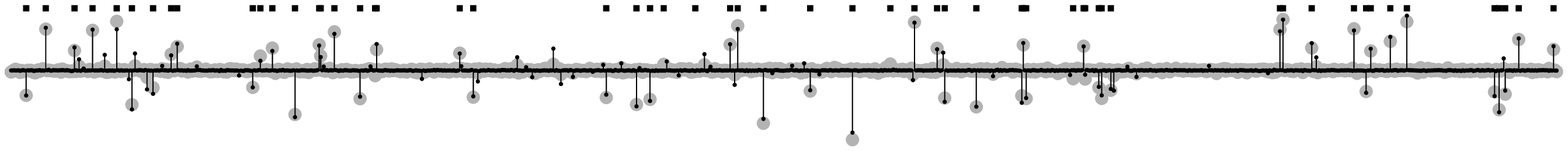}
\includegraphics[width=\textwidth]{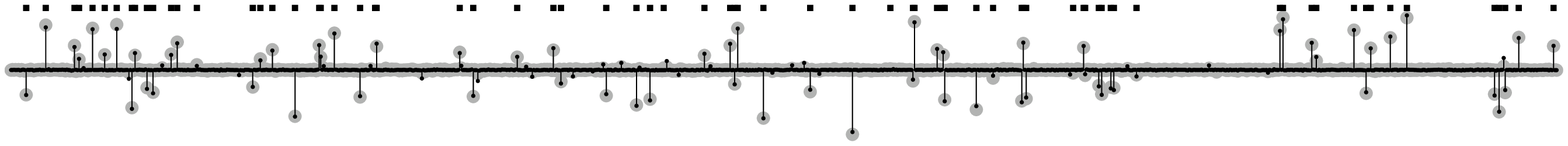}
\includegraphics[width=\textwidth]{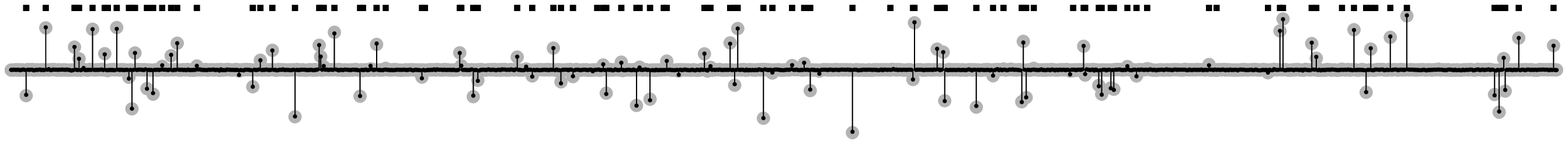}
\includegraphics[width=\textwidth]{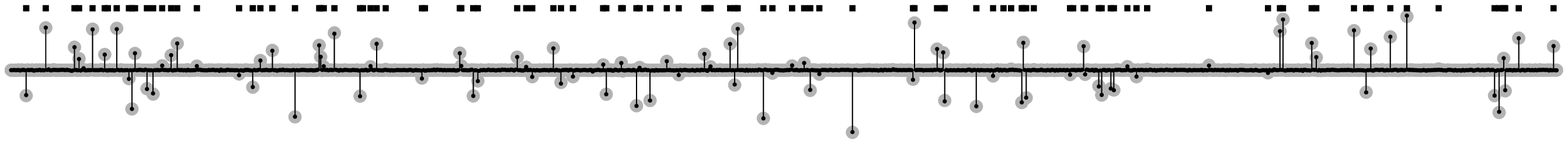}
\caption{Progress of IDE-s toward final solution (Experiment 1) :
$m = 1024$, $n = \lfloor 0.4\,m \rfloor= 409$, $\#act \approx
105$. Each plot shows the original source vector (black) and its
estimate obtained after an iteration (gray). The sources detected
to be active are marked with a black square above each plot. Six
iterations were used with threshold values (form top to bottom)
$\epsilon = 0.3,\,0.2,\,0.1,\,0.05,\,0.02,\,0.01$. The top plot
corresponds to the first iteration. - } \label{fig:idps_progress}
\end{figure*}
}

\newcommand{\figmpvsidpx}{
\begin{figure}[ptb]
\centering
\includegraphics[width=6cm]{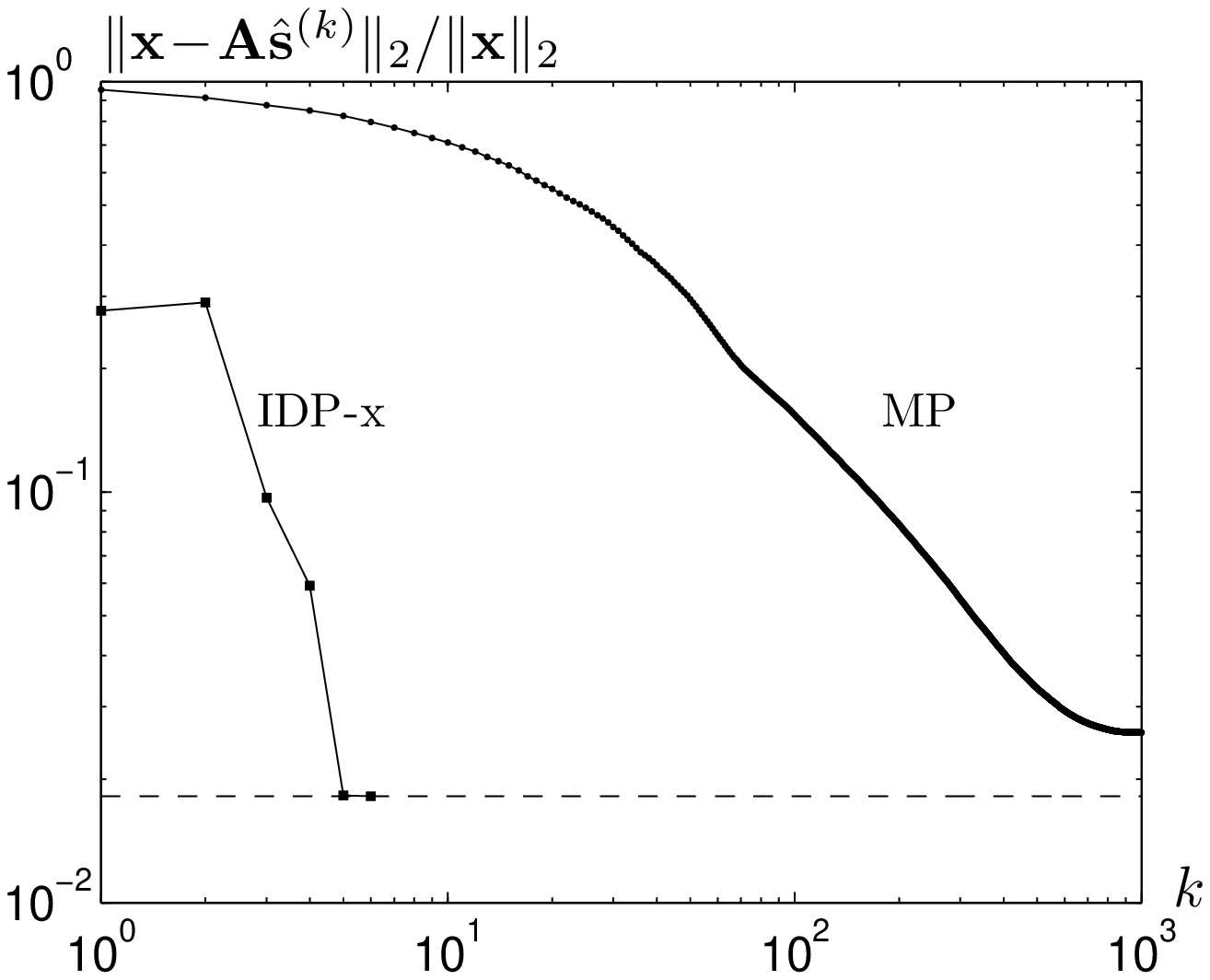}
\caption{IDE-x versus MP: Relative approximation error in $\xb$
obtained by IDE-x/MP at each iteration/step plotted against
iteration/step index ($k$). The data is from experiment
1.}\label{fig:mp_vs_idpx}
\end{figure}
}

\newcommand{\tempoavg}{
\begin{figure}[ptb]
\centering
\includegraphics[width=9cm]{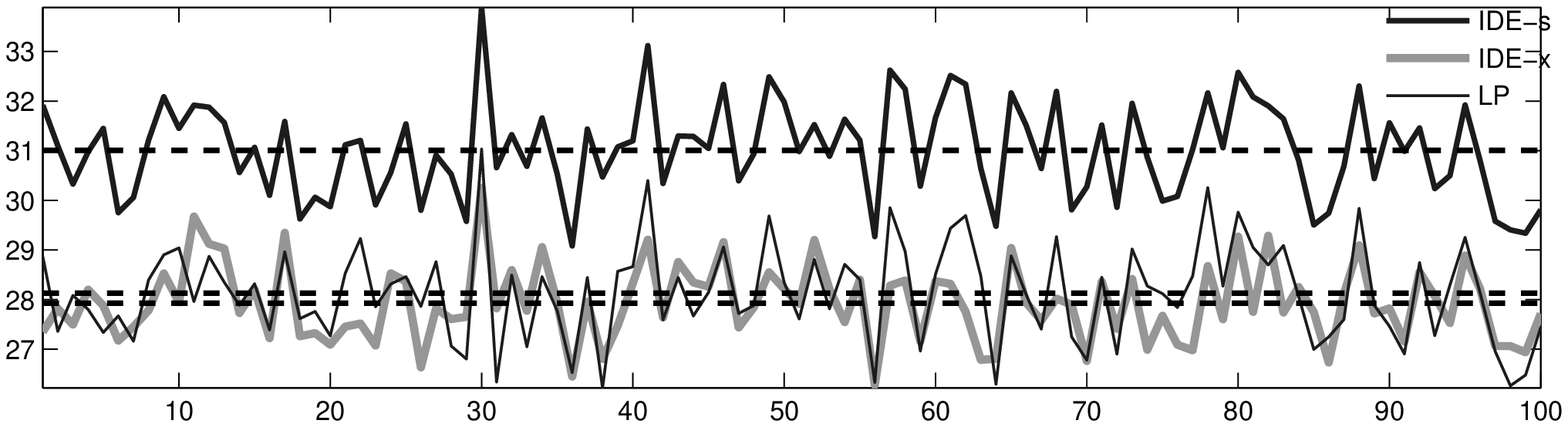}\\
\includegraphics[width=9cm]{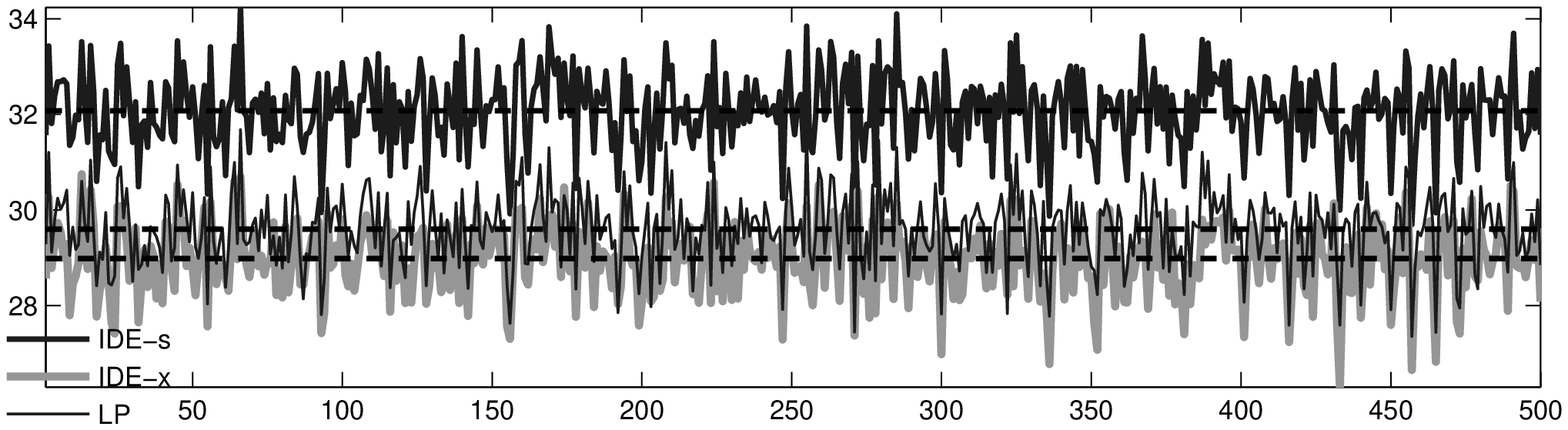}\\
\includegraphics[width=9cm]{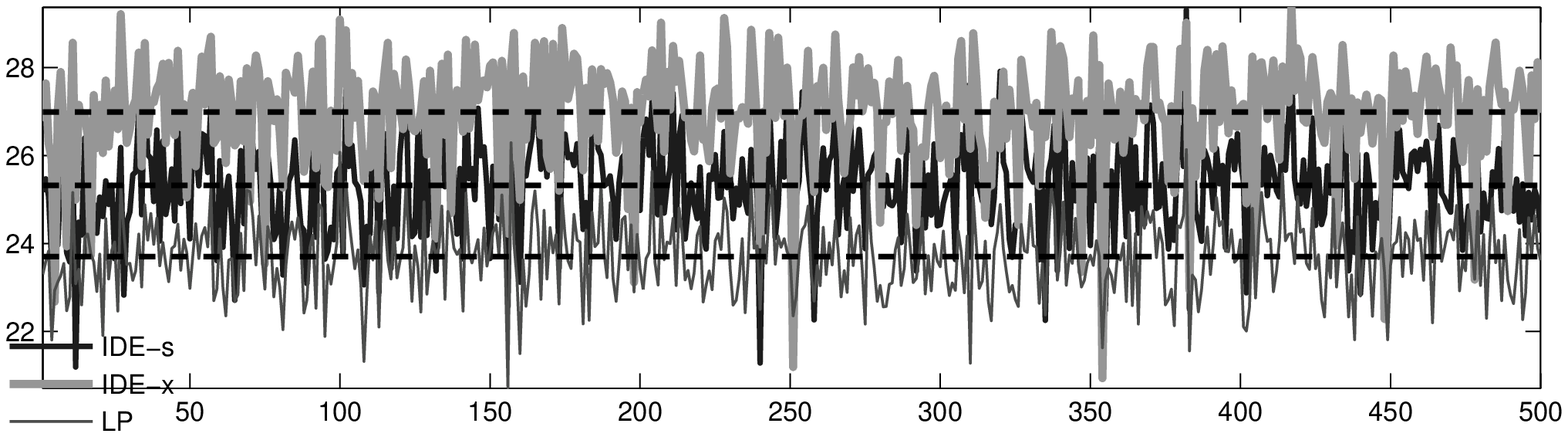}
\caption{Temporal SNR versus the source index ($1 \le i \le m$)
for the three algorithms IDE-s, IDE-x and LP, in three different
settings: (top) $(m,n/m)=(100,0.6)$, (middle) $(m,n/m)=(500,0.6)$
and (bottom) $ (m,n/m)=(500,0.4)$. Temporal averages are over
$N=1000$ samples. All samples are drawn from a Gaussian mixture
model (for the sources) with $\mypi{0} = 0.9$ and
$\sigma_{0}/\sigma_{1} = 0.01$.} \label{fig:tempo_avg}
\end{figure}
}

\newcommand{\speedtest}{
\begin{figure}[ptb]
\centering
\includegraphics[width=6cm]{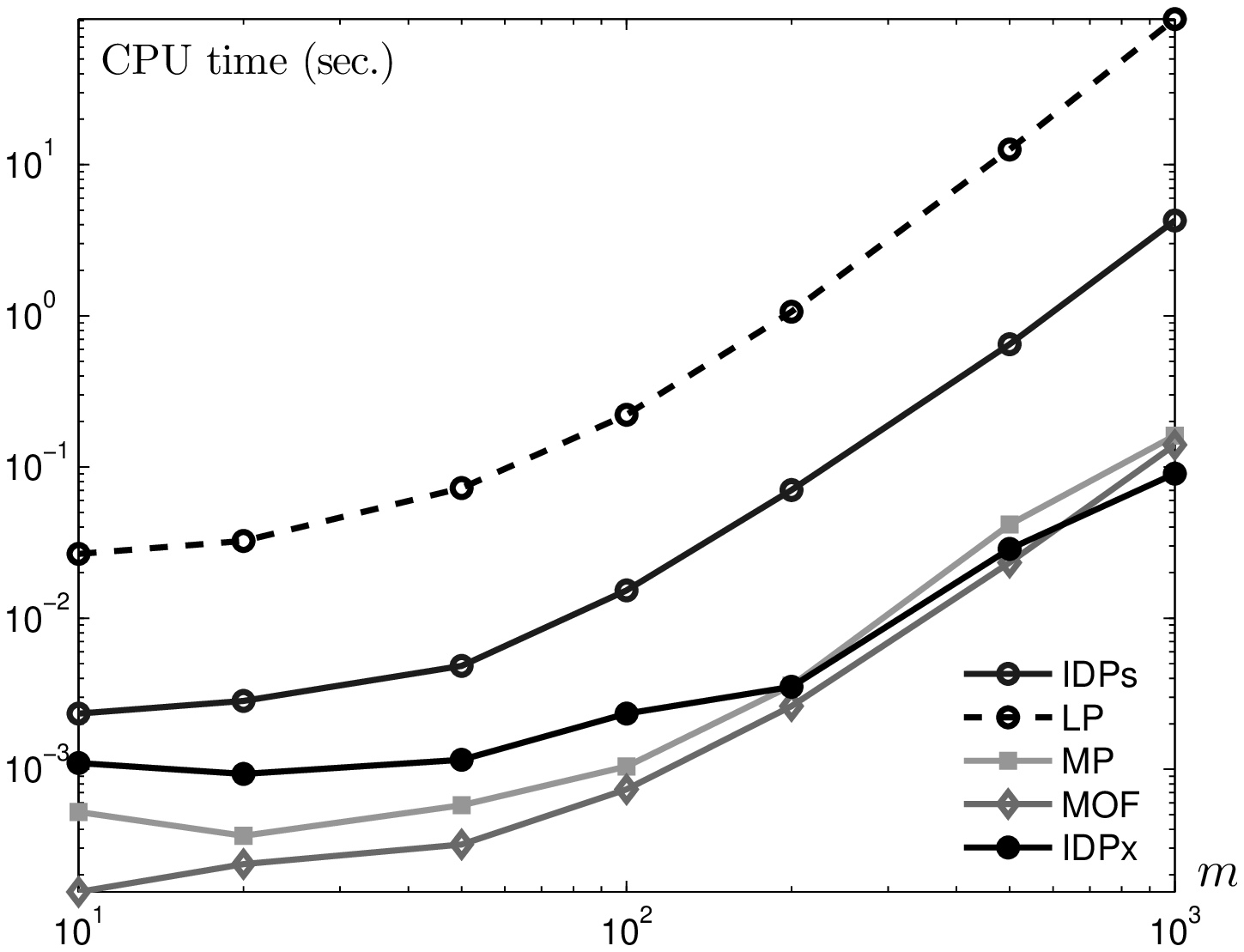}
\caption{Average CPU time in sec. versus problem dimension ($m$)
for various algorithms. At all dimensions, $n=0.4\,m$. Temporal
averages are over N = 10 samples.} \label{fig:speed_test}
\end{figure}
}

\newcommand{\sparsitythresh}{
\begin{figure}[tb]
\centering
 \ifonecol
\includegraphics[width=6cm]{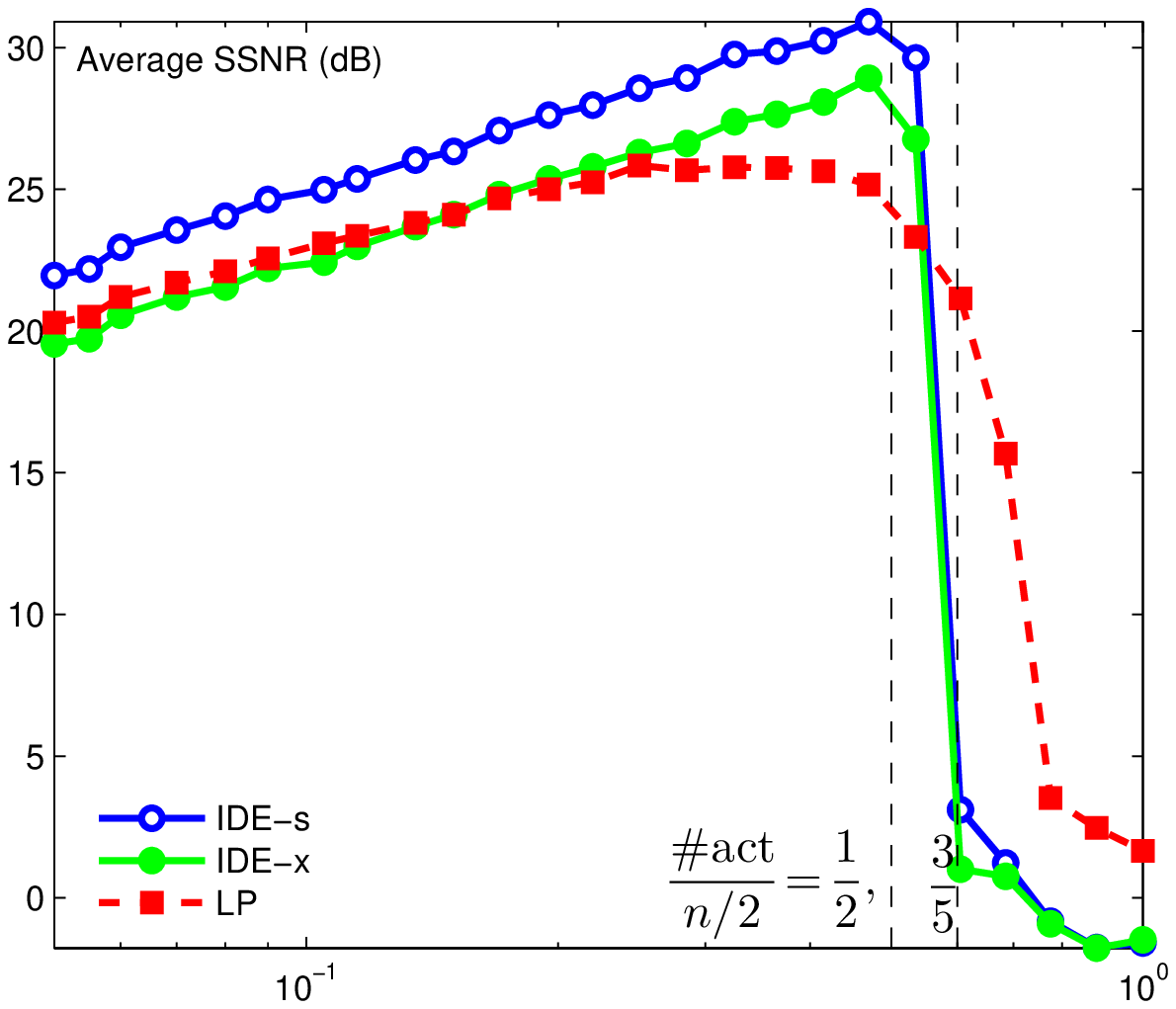}\hfil
\includegraphics[width=6cm]{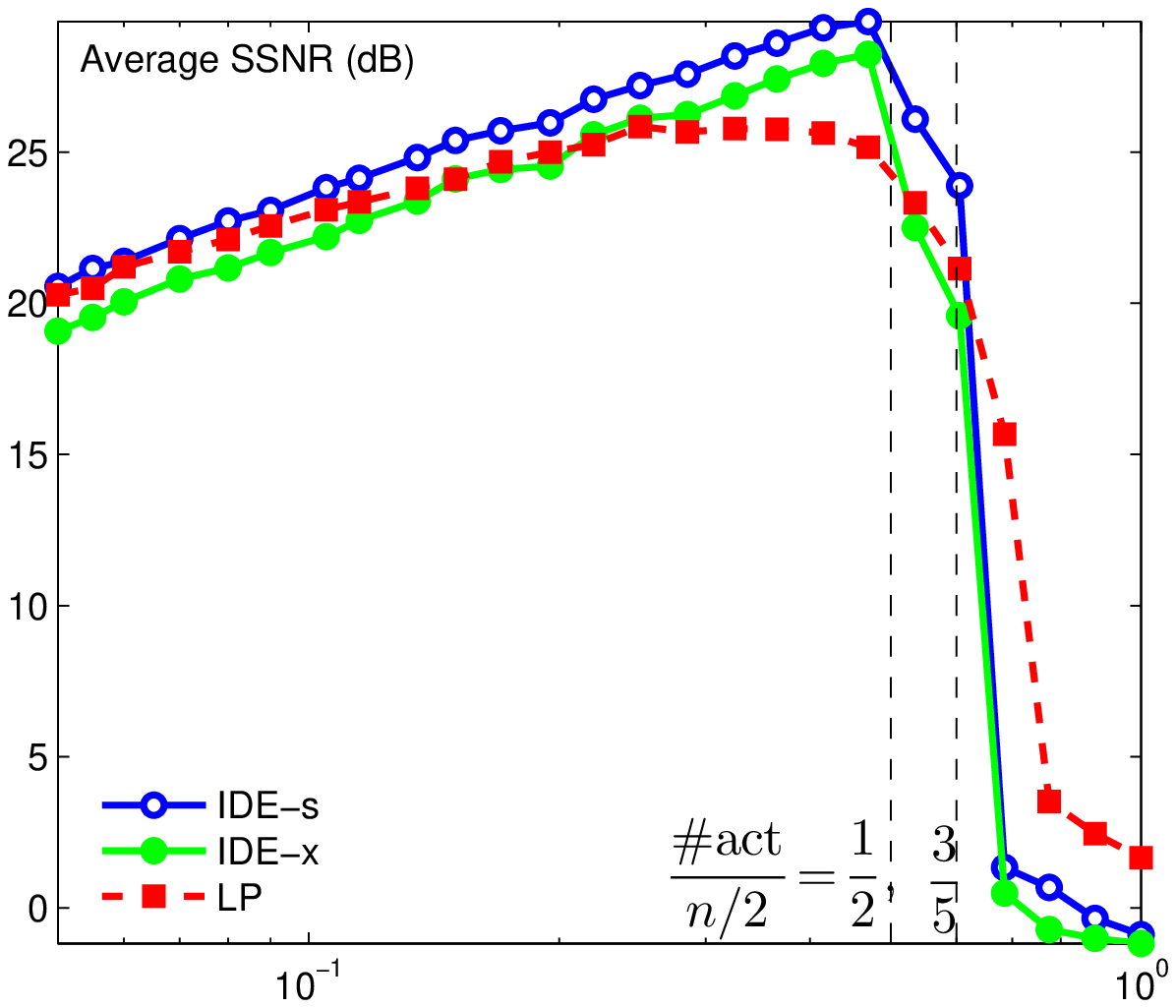}\\
 \fi
{\vspace{-0.25cm}(a) \hfil \hspace{2cm} (b)}\\
\vspace{0.25cm}\caption{Average SSNR (in dB) versus normalized
number of active sources, $\#act/(n/2)$, as a measure of
sparsity. For each value of $\#act$, the average is obtained over
$N=10$ samples. The two plots correspond to different threshold
sequences used in implementing IDEs: (a) 10-point sequence form
experiment 2, (b) a more refined 13-point sequence.}
\label{fig:sparsity_thresh}
\end{figure}
}

\newcommand{\noiseinA}{
\begin{figure}[tb]
\centering
\includegraphics[width=6cm]{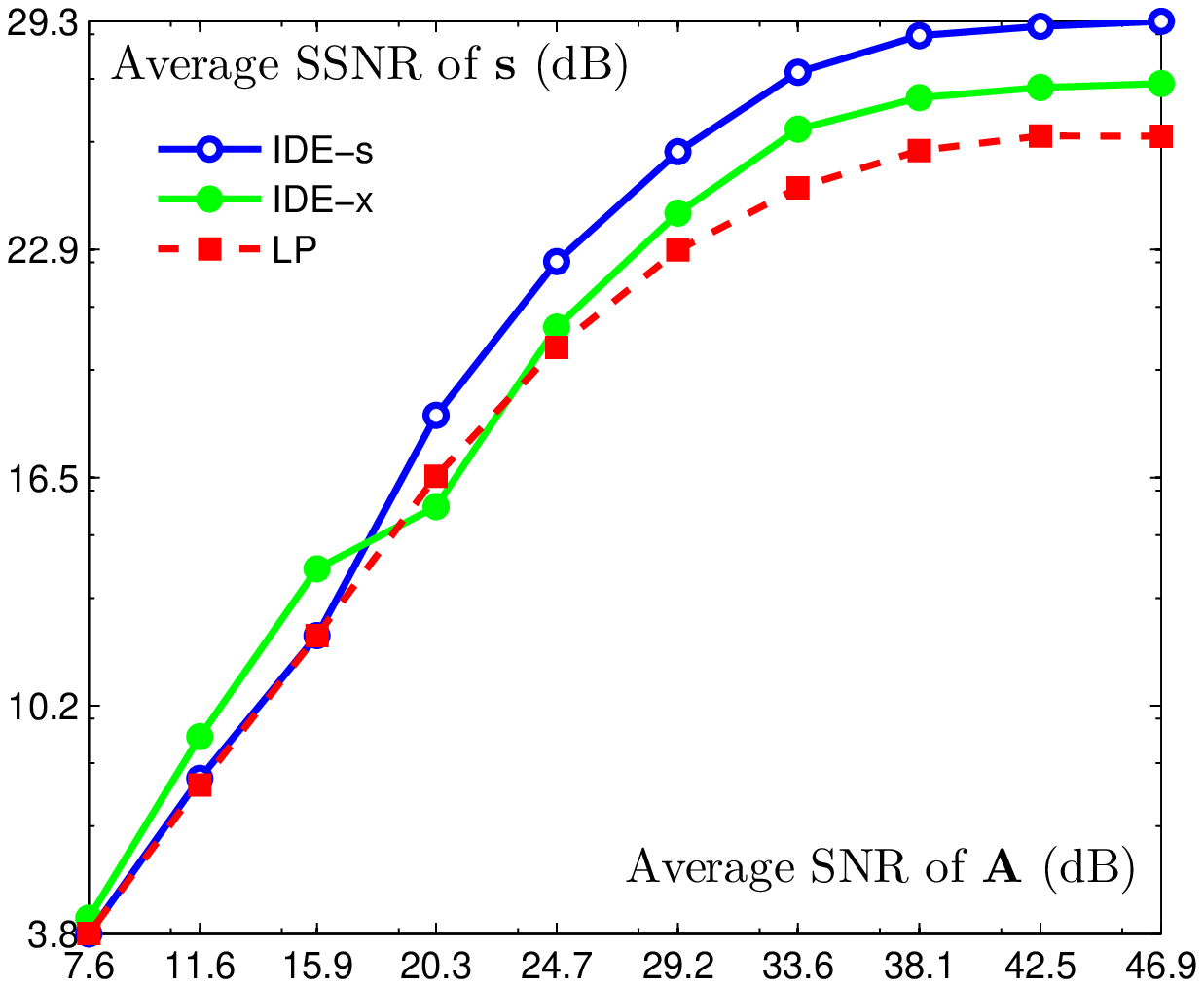}
\caption{Effect of noisy $\Ab$ on performance: plots of average
SSNR (in dB) versus average SNR of matrix $\Ab$. The averages are
obtained over $N=10$ noisy realizations of $\Ab$. The original
$\Ab$ is $200\times500$.} \label{fig:noise_in_A}
\end{figure}
}

\newcommand{\idescheme}{
\begin{figure}[tb]
\centering
\includegraphics[width=8cm]{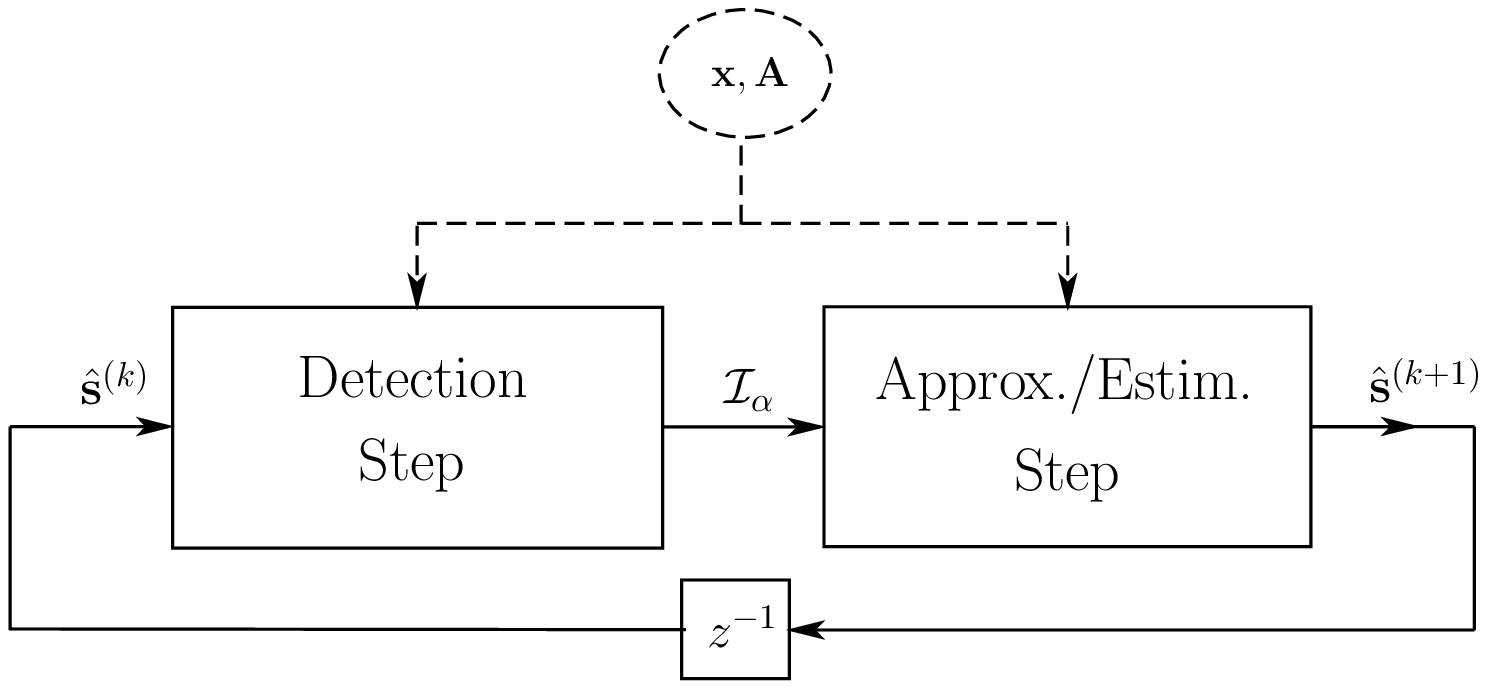}
\caption{Schematic diagram illustrating IDE operation:
$\sbh^{(k)}$ is the source vector estimate after $k$-th iteration;
$\mathcal{I}_{\alpha}$ denotes the (set of) indices of sources
detected active.} \label{fig:IDE_scheme}
\end{figure}
 }